%% file: main.tex
\documentclass[10pt,twocolumn]{article}

\usepackage[T1]{fontenc}
\usepackage[utf8]{inputenc}
\usepackage[english]{babel}
\usepackage{lmodern}
\usepackage{microtype}
\usepackage[margin=0.85in]{geometry}
\usepackage{amsmath,amssymb}
\usepackage{booktabs}
\usepackage{array}
\usepackage{multirow}
\usepackage{graphicx}
\usepackage{xcolor}
\usepackage{caption}
\usepackage{subcaption}
\usepackage{enumitem}
\usepackage[numbers,sort&compress]{natbib}
\usepackage{hyperref}
\hypersetup{colorlinks=true, linkcolor=blue!55!black, citecolor=green!45!black,
  urlcolor=blue!70!black}
\usepackage{cleveref}
\usepackage{tikz}
\usetikzlibrary{arrows.meta,positioning,fit,backgrounds,calc,decorations.pathreplacing}

\definecolor{ink}{HTML}{21303F}
\definecolor{accent}{HTML}{B1283A}
\definecolor{cool}{HTML}{2F6F8F}
\definecolor{nullc}{HTML}{9AA3AB}
\definecolor{gridc}{HTML}{C7CCD1}

\graphicspath{{figs/}}
\newcommand{\ci}[2]{{\scriptsize[#1,\,#2]}}

\usepackage{stfloats}

\title{\vspace{-2.5em}\bfseries Verifiable Rewards for Calibrated Probabilistic Forecasting}
\author{Sadanand Singh \quad Allam Reddy \quad Manan Chopra \\[3pt]
\normalsize Cascade Research}
\date{}

\begin{document}
\twocolumn[
\maketitle
\begin{@twocolumnfalse}
\begin{abstract}
\noindent
Reinforcement learning with verifiable rewards can in principle train calibrated probabilistic
forecasters, since a proper scoring rule such as the Brier score is computed from outcomes alone and is
minimized in expectation by the true probability. In practice it degrades calibration, and existing
remedies address epistemic uncertainty, where a model's confidence accompanies a verifiably correct or
incorrect answer. We study aleatoric forecasting, where the forecast itself is the output and the label
is one stochastic outcome, taking NFL in-game win probability as a testbed with the betting market as a
reference. Rewarding the realized per-play outcome fails, because the single outcome is a noisy target
and the policy gradient corrupts the chain of thought. We introduce a verifiable, label-free reward, a
state-conditioned empirical win rate estimated from past outcomes, that removes the label noise, and we
keep the gradient off the reasoning, by direct prediction or a gradient mask, so it cannot be corrupted.
Trained with this reward alone, without human labels or supervised fine-tuning, a 7B model reaches the
calibration of the betting market by direct prediction and is better calibrated than a zero-shot
frontier model. That frontier model and a tabular estimator reach the same Brier score as this model,
identifying the market's small remaining edge as live in-game information beyond their shared
inputs. Masking the gradient,
rather than dropping the chain of thought, preserves reasoning from which the forecast follows, which
ordinary chain-of-thought training corrupts.
\end{abstract}
\vspace{1.2em}
\end{@twocolumnfalse}
]

\input{sections/01_intro.tex}

\input{sections/08_related.tex}

\input{sections/02_setting.tex}

\section{Method}\label{sec:method}
We post-train the forecaster with group-relative reinforcement learning, computing the reward from a
state-conditioned estimate of the win rate rather than from the realized outcome, and restricting the
policy gradient to the tokens that carry the answer rather than to the whole completion.

\input{sections/03_reward.tex}

\input{sections/04_decoupling.tex}

\input{sections/05_setup.tex}

\input{sections/06_results.tex}

\input{sections/09_conclusion.tex}

\bibliographystyle{unsrtnat}
\bibliography{references}

\appendix
\onecolumn
\input{sections/10_appendix.tex}

\end{document}

%% file: sections/01_intro.tex
\section{Introduction}
\label{sec:intro}

\input{sections/F1_principle.tex}

Reinforcement learning with verifiable rewards (RLVR) post-trains a language model against a reward
computed from observed outcomes \citep{grpo, deepseek_r1}. Calibrated probabilistic prediction is a
natural target for it: a proper scoring rule such as the Brier score is computable from outcomes alone
and is minimized in expectation by the true probability \citep{brier1950, gneiting2007}, so optimizing
it should produce forecasts whose probabilities match observed frequencies.

In practice, reinforcement learning does the opposite, degrading calibration and leaving models
overconfident, whether the reward is human feedback \citep{leng2024} or a verifiable correctness signal
\citep{dcpo, uncalibrated_reasoning}. The work that corrects this addresses epistemic uncertainty, where
the model answers a question that has a correct answer and the goal is a calibrated confidence that the
answer is right \citep{rlcr, dcpo}.

A second kind of uncertainty is left untreated. It is aleatoric: the output is itself a probability, and
the label is a single realized outcome of a stochastic event, with no answer that can be called correct.
NFL in-game win probability is a clear instance. At any point in a game the forecaster states the
probability that the team in possession wins; the realized result is one Bernoulli draw from a rate that
no model observes directly; and the betting market provides a strong, independent estimate of that rate.
Whether RLVR can train a calibrated forecaster in this regime, and what makes it fail, has not been
studied.

Reinforcement learning against the per-play Brier score decalibrates such a forecaster through two
mechanisms. The label for a play is a single realized outcome, one draw from the rate being estimated,
so a reward scored against it has high variance and pulls the policy toward whichever result occurred.
When the model reasons in language before answering, a second problem appears: optimizing the final
probability rewrites the reasoning into incoherent arguments. We remove the variance by rewarding a
state-conditioned empirical win rate estimated from past outcomes, a verifiable and label-free target,
and we remove the corruption by keeping the policy gradient off the reasoning, through direct prediction
or a gradient mask over the answer tokens (\cref{fig:principle}). We train Qwen2.5-7B-Instruct without supervised fine-tuning
and evaluate on held-out seasons against the betting market, a frontier model, and the empirical-rate
table.

Trained by direct prediction, the model matches the market's calibration on held-out games, at an
expected calibration error (ECE) of $0.029$ against the market's $0.027$, and it is better calibrated
than a zero-shot frontier model, with no human labels and no access to the market. On the Brier score it
matches that frontier model and a tabular estimator, the limit that the public game state allows; the
market's small remaining edge is the live in-game information none of them can see. Masking
the gradient instead of discarding the reasoning keeps the chain of thought faithful, lowering the rate
at which the stated probability does not follow from it from $22.4\%$ to $4.4\%$, at a small cost in
sharpness.

\paragraph{Contributions.}
\begin{enumerate}[leftmargin=1.4em,itemsep=2pt,topsep=2pt]
\item Our reward, a state-conditioned empirical win rate estimated from realized outcomes, is verifiable
and label-free, and replaces the single realized outcome, whose variance decalibrates per-play Brier
training; on held-out games it scores within $0.007$ Brier of the betting market
(\cref{sec:reward}).
\item Reinforcement learning decalibrates the forecaster through the gradient on the chain of thought,
not through the reward: with the rate target, training the full completion drives held-out ECE
from $0.19$ to $0.30$, whereas confining the gradient to the answer, by direct prediction or an
answer-span mask, holds it near the market's (\cref{sec:decoupling}).
\item With the chain of thought dropped, the model reaches the information ceiling of the public game
state: it matches the market's calibration ($0.029$ against $0.027$ ECE), is better calibrated than a
zero-shot frontier model, and
reaches the same Brier score as that frontier model and a tabular rate estimator, leaving the market's
edge as live in-game information (\cref{sec:results}).
\end{enumerate}

\noindent Code, the prepared data, the per-play predictions, and the trained adapters are publicly
available.\footnote{Code: \url{https://github.com/jasper-research/nfl-rlvr-release}. Data and trained
adapters: \url{https://doi.org/10.5281/zenodo.21082572}.}

%% file: sections/F1_principle.tex
\begin{figure*}[t]
\centering
\resizebox{\linewidth}{!}{%
\begin{tikzpicture}[
  >={Stealth[length=2mm]},
  font=\small,
  rowlbl/.style={anchor=west, font=\bfseries\small},
  sub/.style={anchor=north west, font=\scriptsize\itshape, align=left, text width=30mm},
  reason/.style={rounded corners=2pt, draw=ink!65, fill=ink!4, align=center,
    text width=23mm, minimum height=10mm, inner sep=2pt, font=\footnotesize},
  ans/.style={rounded corners=2pt, draw=ink!65, align=center,
    text width=16mm, minimum height=10mm, inner sep=2pt, font=\footnotesize\bfseries},
  inp/.style={rounded corners=2pt, draw=ink!65, fill=ink!7, align=center,
    text width=20mm, minimum height=12mm, inner sep=3pt, font=\footnotesize},
  tgt/.style={rounded corners=2pt, align=center, text width=27mm,
    minimum height=8mm, inner sep=2pt, font=\footnotesize},
  flow/.style={->, semithick, ink!75},
]

\node[inp] (S) at (2.6,0) {Game state $x$\\\scriptsize score, time, spread};

\def\ty{2.05}
\node[rowlbl, text=cool] at (-0.2,\ty+0.82) {Naive};
\node[sub, text=cool!80!black] at (-0.2,\ty+0.52) {reward the outcome,\\train the whole\\completion};

\node[reason] (TR) at (5.7,\ty) {chain of thought};
\node[ans, draw=cool!70, fill=cool!14] (TA) at (7.95,\ty) {P\,=\,88\%};
\begin{scope}[on background layer]
  \node[rounded corners=3pt, draw=cool!55, fill=cool!5, fit=(TR)(TA), inner sep=2.5pt] (TC) {};
\end{scope}
\draw[flow] (S) to[out=70,in=180] (TR.west);

\node[tgt, draw=cool!60, fill=cool!8] (TT) at (7.6,\ty+1.95)
  {single outcome $y\in\{0,1\}$\\\scriptsize one noisy Bernoulli draw};
\draw[flow, cool!85] (TA.north) -- node[right, font=\scriptsize, pos=0.45] {$r=1-(p-y)^2$} (TT.south);

\draw[decorate, decoration={brace, amplitude=4pt, mirror}, cool!85, semithick]
  (TC.south west) -- (TC.south east);
\node[font=\scriptsize, text=cool!80!black, align=center, text width=42mm]
  at (6.85,\ty-1.25) {policy gradient over the whole completion\\\textit{reasoning is corrupted}};

\begin{scope}[shift={(10.85,\ty-0.67)}, scale=1.35]
  \draw[gridc, thin] (0,0) rectangle (1,1);
  \draw[dashed, ink!45, line width=0.4pt] (0,0) -- (1,1);
  \draw[cool, semithick] (0.08,0.34) -- (0.92,0.66);
  \foreach \x/\y in {0.18/0.38,0.4/0.46,0.6/0.54,0.82/0.628} \fill[cool] (\x,\y) circle (0.032);
\end{scope}
\node[font=\scriptsize, text=cool!80!black, align=center] at (11.55,\ty-0.95) {overconfident};

\def\by{-2.05}
\node[rowlbl, text=accent] at (-0.2,\by+0.82) {Ours};
\node[sub, text=accent!82!black] at (-0.2,\by+0.52) {reward the rate,\\train the answer span};

\node[reason] (BR) at (5.7,\by) {chain of thought};
\node[ans, draw=accent!75, fill=accent!12] (BA) at (7.95,\by) {P\,=\,61\%};
\begin{scope}[on background layer]
  \node[rounded corners=3pt, draw=ink!30, fill=ink!2, fit=(BR)(BA), inner sep=2.5pt] (BC) {};
\end{scope}
\draw[flow] (S) to[out=-70,in=180] (BR.west);

\node[tgt, draw=accent!70, fill=accent!7] (BT) at (7.6,\by-1.95)
  {empirical rate $\hat p(x)$\\\scriptsize win rate over similar states};
\draw[flow, accent!85] (BA.south) -- node[right, font=\scriptsize, pos=0.45] {$r=1-(p-\hat p)^2$} (BT.north);

\draw[decorate, decoration={brace, amplitude=4pt}, accent!85, semithick]
  (BA.north west) -- (BA.north east);
\node[font=\scriptsize, text=accent!82!black, align=center, text width=44mm]
  at (7.95,\by+1.2) {gradient on the answer span only\\\textit{reasoning is preserved}};

\begin{scope}[shift={(10.85,\by-0.67)}, scale=1.35]
  \draw[gridc, thin] (0,0) rectangle (1,1);
  \draw[dashed, ink!45, line width=0.4pt] (0,0) -- (1,1);
  \draw[accent, semithick] (0.06,0.06) -- (0.94,0.94);
  \foreach \x in {0.18,0.4,0.6,0.82} \fill[accent] (\x,\x) circle (0.032);
\end{scope}
\node[font=\scriptsize, text=accent!82!black, align=center] at (11.55,\by-0.95)
  {calibrated to\\the market};

\end{tikzpicture}%
}
\caption{The same game state, two ways to train a 7B model against outcomes. Rewarding the single
realized outcome and updating the whole completion drives the chain of thought to extreme, overconfident
numbers. Rewarding a state-conditioned empirical win rate $\hat p(x)$ and confining the gradient to the
answer span leaves the reasoning intact and calibrates the forecaster to the betting market, using no
human labels and no supervised fine-tuning.}
\label{fig:principle}
\end{figure*}
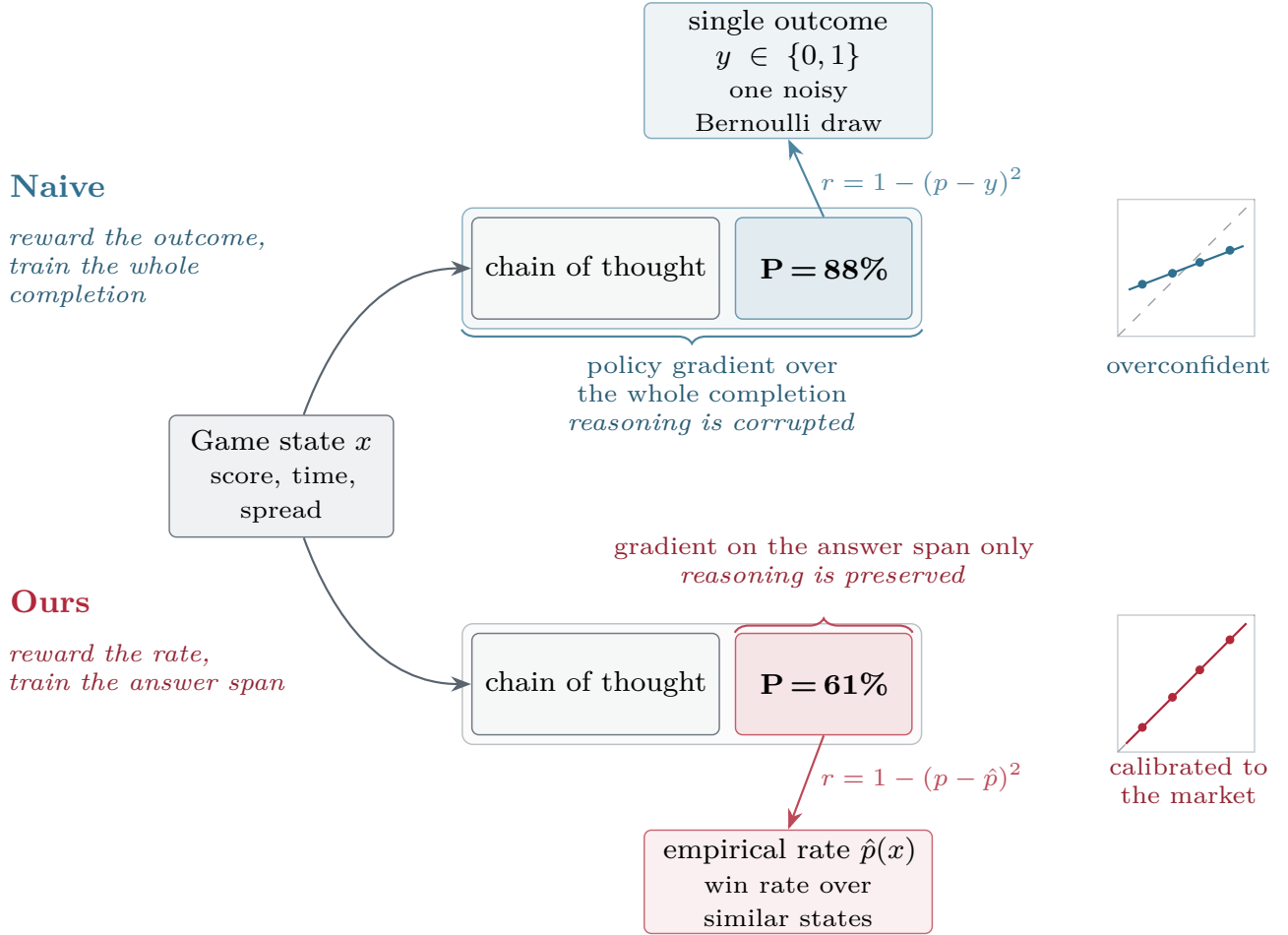

%% file: sections/08_related.tex
\section{Related work}
\label{sec:related}

Reinforcement learning improves accuracy but degrades calibration, leaving models overconfident.
\citet{leng2024} attribute this, under human feedback, to a reward model that prefers confident answers.
\citet{dcpo} identify a conflict between the accuracy and calibration gradients under verifiable rewards,
and decouple the two with a masked-gradient update. \citet{uncalibrated_reasoning} attribute the
overconfidence on binary stochastic outcomes to the standard-deviation normalization in the
group-relative advantage, which they remove. These analyses take the realized outcome as the target.
With a denoised conditional-rate target we find instead that the normalization is necessary, and that the
overconfidence is driven by the gradient on the reasoning rather than by the advantage.

A second class of methods rewards a proper scoring rule on a verbalized confidence. \citet{rlcr} add a
Brier-score term to a correctness reward, \citet{rewarding_doubt} use the logarithmic score, and
\citet{band2024} reward a downstream reader's accuracy. These score a confidence in an answer that is
right or wrong. We score the forecast itself against a state-conditioned empirical rate, where the label
is a single stochastic outcome and correctness is undefined.

Language models are also trained and evaluated as event forecasters. \citet{halawi2024} approach
human-crowd accuracy through retrieval and aggregation, and \citet{pratt2024} find that prompting
strategies do not yield calibrated forecasts. \citet{turtel2025outcome} reward realized outcomes on
prediction-market and news questions, and \citet{turtel2025selftaught} rank self-generated forecasts
against outcomes. Both improve calibration while keeping the chain of thought and rewarding the outcome,
whereas we reward the conditional rate and mask the gradient off the reasoning. \citet{paleka2025} and
\citet{karger2024forecastbench} document temporal leakage in such evaluations, which our season-disjoint
test on resolved games avoids.

Our training is a choice of credit assignment within group-relative optimization \citep{grpo}. DAPO
\citep{dapo} and GPG \citep{gpg2026} remove the KL penalty and reference model, as we do. \citet{rpg2026}
analyze the instability of the KL estimator when the gradient concentrates on a few tokens.
\citet{wang2025entropy} restrict updates to high-entropy tokens, and \citet{archer2025, gtpo2025} treat
reasoning and answer tokens differently. Our answer-span mask restricts the gradient by token role.

Win-probability estimation for the NFL ranges from random forests \citep{lock2014} to the boosted models
of the nflverse pipeline \citep{yurko2018, nflfastr}, which we use as static baselines. \citet{brill2024}
show that play-by-play data bounds the accuracy of any such model, consistent with our reading of the
market's edge as structural. \citet{polson2015} treat the point spread as an implied volatility. The
betting market is known to predict outcomes more accurately than statistical models or individual bettors
\citep{boulier2003, levitt2004, franck2010, cox2021}. We convert market odds to probabilities following
\citet{strumbelj2014} and use them only as a reference.

%% file: sections/02_setting.tex
\section{Setting and background}
\label{sec:setting}

We forecast in-game win probability in American football. The state at a given play comprises the score
margin, the quarter and time remaining, down and distance, field position, the team in possession, and
the public pregame point spread. The forecaster returns a single probability that the team in possession
wins. We draw states from regular-season National Football League games and partition them by season,
training on 2015 through 2022, selecting on 2023, and testing on 2024, so that no game appears in more
than one split.
The only market signal in the prompt is the pregame point spread, which is fixed before kickoff and
public; the live win probability the market quotes during the game is withheld from both the model and
the reward, and enters only at evaluation.

The output is a probability, and the label is a single realized outcome of a stochastic event. This
separates the task from the calibration problems usually studied in reinforcement learning, where the
model answers a question that has a correct answer and the quantity of interest is its confidence that
the answer is right \citep{rlcr, dcpo}. That uncertainty is epistemic, and a stronger model can in
principle reduce it. The uncertainty here is aleatoric: the outcome is random given the state, and no
forecaster can remove it. The target of calibration is therefore the conditional win rate
$\eta(x)=\Pr(\text{win}\mid x)$, not the correctness of an answer.

We measure a forecaster by how closely its probabilities match observed frequencies. The Brier score,
the mean squared error $(p-y)^2$ between a forecast $p$ and an outcome $y\in\{0,1\}$ \citep{brier1950},
is strictly proper: among all functions of the state, its expectation is minimized uniquely by $\eta(x)$
\citep{gneiting2007}. It decomposes into reliability, resolution, and uncertainty \citep{murphy1973}.
Reliability is calibration, the agreement between stated probabilities and realized frequencies;
resolution is sharpness, the spread of forecasts across states of differing outcome rate; uncertainty is
fixed by the base rate. Because two forecasters can be equally calibrated yet differ in Brier score
through resolution alone, we report calibration on its own, through the expected and maximum calibration
error and reliability diagrams \citep{guo2017, brocker2007, dimitriadis2021}. We compare forecasters with
a paired bootstrap over plays \citep{efron1979}. Calibration error also tests whether training found
$\eta$ rather than memorizing the training outcomes, since a model that memorized them would be
miscalibrated on held-out games.

We take the betting market as the ceiling: it quotes a live win probability for each state and predicts
outcomes more accurately than statistical models built from game features \citep{boulier2003,
levitt2004}. We convert its odds to a probability \citep{strumbelj2014}. The gap between a forecaster and
the market then measures the live in-game information that the public state lacks.

%% file: sections/03_reward.tex
\subsection{Conditional-rate reward}
\label{sec:reward}

The ideal reward would score each forecast against the true rate $\eta(x)$, but $\eta(x)$ is never
observed. The only label for a play is its realized outcome $y$, a single draw from that rate. A reward
built on the draw, $r=1-(p-y)^2$, is unbiased but noisy, since two identical favorites, one that wins
and one that loses, produce equal and opposite gradients. The noise does not average out with more plays
at a fixed state, because it is intrinsic to a binary label, and in the small groups that GRPO compares
\citep{grpo} it overwhelms the per-state signal.

We score each forecast against an estimate of the rate instead. We bin the training plays by score
margin, time remaining, and pregame point spread, and set each bin to the fraction of its plays the
possession team won. Bins with few plays are shrunk toward the overall rate by an empirical-Bayes prior,
so a sparse bin is not governed by its few plays \citep{efron_morris1975}. The estimate
$\hat p(x)$ uses only realized outcomes and the public pregame line, never the live market probability,
which we hold out for evaluation. The reward is
\begin{equation}
  r \;=\; 1 - \bigl(p - \hat p(x)\bigr)^2 .
  \label{eq:reward}
\end{equation}
The target uses no human labels. It comes from the same outcomes a binary reward would use, averaged
over states to yield a low-variance estimate of $\eta(x)$. On held-out 2024 games it scores a Brier of
$0.143$, close to the market's $0.136$, and it supplies a dense reward at every state. We call $\hat p$
the teacher, because a model trained on \cref{eq:reward} can be no better calibrated than $\hat p$
itself.

An ablation under identical training isolates the target as the cause (\cref{tab:reward}). Rewarding the
realized outcome reaches a Brier of $0.166$, but its calibration error drifts to $0.10$ as the policy
fits individual outcomes. An equal blend of outcome and rate is worse on both, $0.181$ and $0.121$,
because it reintroduces the outcome's noise. The rate alone is best on both, at $0.154$ and $0.050$.

\begin{table}[t]
\centering
\input{tables/T2_reward.tex}
\caption{Reward-target ablation under identical training (direct prediction, \cref{sec:decoupling};
in-training held-out, $n=128$). The realized-outcome reward is higher-variance and its calibration
drifts; blending the outcome back in is worse on both axes; the conditional rate $\hat p$ is the most
accurate and the most stable.}
\label{tab:reward}
\end{table}

%% file: tables/T2_reward.tex
\begin{tabular}{lcc}
\toprule
Reward target & Brier & ECE \\
\midrule
Realized outcome $y\in\{0,1\}$ & 0.166 & 0.10 \\
Blend $\tfrac{1}{2}y + \tfrac{1}{2}\hat{p}$ & 0.181 & 0.121 \\
Empirical rate $\hat{p}$ (ours) & 0.154 & 0.050 \\
\bottomrule
\end{tabular}

%% file: sections/04_decoupling.tex
\subsection{Decoupling reasoning from the gradient}
\label{sec:decoupling}

\begin{figure*}[t]
\centering
\includegraphics[width=\linewidth]{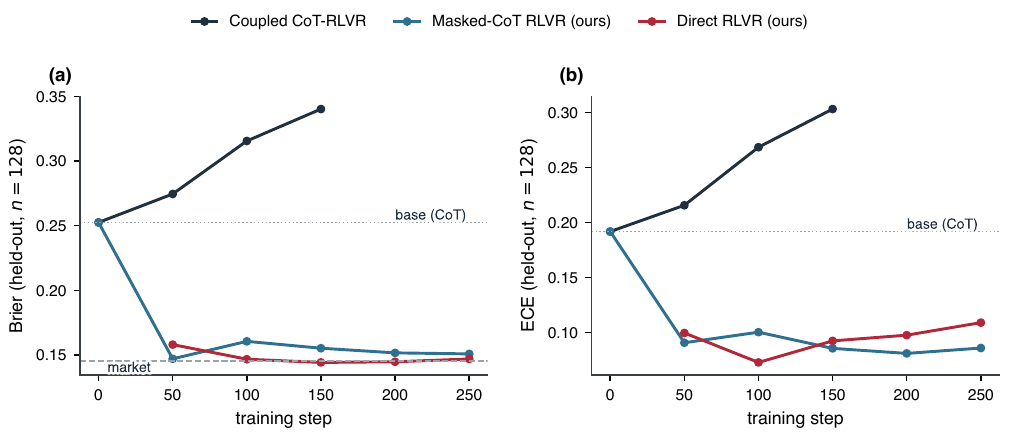}
\caption{Held-out calibration during training (fixed $n=128$). Applied to the full completion, the
reward of \cref{eq:reward} drives a chain-of-thought policy away from calibration on both Brier (a) and
ECE (b). Removing the reasoning from the gradient, by direct prediction or by masking the gradient to
the answer span, recovers calibration to the teacher's level. The full-completion and masked runs share
the same reward, prompt, and base model, and differ only in the gradient mask.}
\label{fig:trajectory}
\end{figure*}

The rate target alone is not enough, because a chain of thought before the answer lets the same reward
make calibration worse. Applied to the full completion, the reward of \cref{eq:reward} drives the
held-out Brier score from the base model's $0.25$ to $0.34$ and the calibration error from $0.19$ to
$0.30$ (\cref{fig:trajectory}). The policy gradient passes through the reasoning tokens, so raising the
reward on the final probability rewrites the reasoning before it into pseudo-quantitative arguments for
ever more extreme numbers. This is the accuracy-calibration gradient conflict reported for
verifiable-reward training \citep{dcpo}.

Two changes keep the reward and stop the gradient from reaching the reasoning. The first, the direct
model, drops the chain of thought and emits the probability directly. With no reasoning tokens to
corrupt, the reward calibrates the model, and the held-out Brier score reaches the teacher's level
within fifty steps and stays there. The second, the masked model, keeps the chain of thought but masks
the gradient. We narrow the completion mask, which selects the tokens that receive advantage, to the
final ``Probability: NN\%'' span, so the reasoning is still sampled and still conditions the answer but
gets no gradient. Masking the gradient
needs the KL penalty turned off, because on the few tokens of the answer span the gradient is
concentrated enough to make the penalty's estimator overflow and the loss diverge, and the masked
reasoning no longer needs the reference-model anchor. With the mask in place, the masked model
calibrates as well as the direct model.

The masked model is slightly less accurate than the direct model. On held-out 2024 games the direct
model reaches a Brier of $0.144$ and the masked model $0.152$, a gap of $0.008$ by a paired bootstrap
over plays (95\% CI $[0.006, 0.010]$). The gap is small, and far below the base-to-teacher gap that the
reward closes. Both trained models trail the market on Brier, but the shortfall is in resolution, not
calibration: the direct model's calibration error of $0.029$ matches the market's $0.027$, and its Brier
is higher only because the static prompt cannot see the live information the market uses to separate
states more sharply (\cref{sec:results}).

\begin{table}[t]
\centering
\setlength{\tabcolsep}{4.5pt}
\input{tables/T3_decoupling.tex}
\caption{Paired bootstrap over held-out 2024 plays ($n=5185$, $10^4$ resamples). Masking the gradient
to the answer span costs a small but significant amount of Brier relative to discarding the reasoning;
both decoupled models trail the market by a similar, significant margin that \cref{sec:results}
attributes to resolution rather than calibration.}
\label{tab:decoupling}
\end{table}

The masked model keeps its reasoning coherent. A blinded judge read each completion and decided whether
the stated probability follows from the reasoning, and the rate of inconsistent completions falls from
$22.4\%$ for the base model to $4.4\%$ after masked training. Full-completion reinforcement learning
leaves the base rate unchanged, so the gain comes from the masked objective and not from reinforcement
learning in general, and a control that applies the masked prompt without training accounts for only
part of it. The base model writes arithmetic that does not sum and then reports an unrelated number, or
rates a clear favorite below an even chance, while the masked model reads the live state and states a
probability that follows from it (\cref{fig:reasoning-examples}). A probability lookup cannot do this,
so the masked model is the better choice when a forecast must be explained, at the cost of slightly
lower sharpness than direct prediction.

\begin{figure*}[t]
\centering
\includegraphics[width=\linewidth]{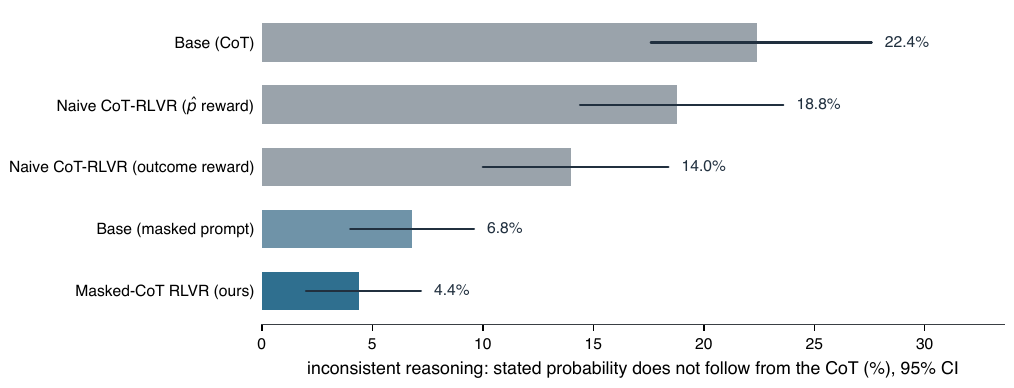}
\caption{Rate at which the stated probability does not follow from the model's own chain of thought, on
a blinded-judge sample of held-out plays (95\% bootstrap intervals). Full-completion chain-of-thought
reinforcement learning leaves the base rate unchanged; masked training reduces it sharply, beyond what
the masked prompt format alone accounts for.}
\label{fig:faithfulness}
\end{figure*}

\input{examples/E1_reasoning.tex}

%% file: tables/T3_decoupling.tex
\begin{tabular}{lcc}
\toprule
Paired comparison (Brier) & $\Delta$ & 95\% CI \\
\midrule
Direct $-$ Masked-CoT & -0.0079$^{\ast}$ & {\scriptsize[-0.0101,\,-0.0057]} \\
Direct $-$ Market & +0.0088$^{\ast}$ & {\scriptsize[0.0068,\,0.0107]} \\
Masked-CoT $-$ Market & +0.0167$^{\ast}$ & {\scriptsize[0.0140,\,0.0194]} \\
Direct $-$ DeepSeek-V4 & +0.0005 & {\scriptsize[-0.0019,\,0.0028]} \\
DeepSeek-V4 $-$ Market & +0.0083$^{\ast}$ & {\scriptsize[0.0064,\,0.0102]} \\
\bottomrule
\end{tabular}

%% file: examples/E1_reasoning.tex
\providecommand{\statehead}[1]{\smallskip\noindent\textbf{#1}\par\smallskip}

\begin{figure*}[t]
\centering
\small
\begin{minipage}{0.99\textwidth}

\statehead{Example 1.\quad Q4, 2:21 left. WAS trailing by 12, 2nd \& 6 at the NYG 8-yard line.
Pregame line: WAS favored by 7.5. Empirical rate $\hat{p}=0.07$; WAS lost.}

\begin{tabular}{@{}p{0.085\textwidth}@{\hskip 6pt}p{0.80\textwidth}@{\hskip 6pt}r@{}}
\textit{Base (CoT)} &
\emph{``[\,\dots] Touchdown and Extra Point: $0.8\times0.95=0.76$ or 76\%. Touchdown and
Two-Point Conversion: $0.8\times0.4=0.32$ or 32\%. Total Probability: $0.76+0.32=1.08$, but
since we can't exceed 100\%, we cap it at 100\%. [\,\dots] approximately 32\%.''} & $0.32$ \\[3pt]
\textit{Masked-CoT (ours)} &
\emph{``[\,\dots] a 12-point deficit with only 2 minutes and 21 seconds left [\,\dots] makes a
comeback extremely challenging. Probability: 8\%.''} & $0.08$ \\
\end{tabular}

\medskip\hrule\medskip

\statehead{Example 2.\quad Q4, 7:57 left. CHI leading by 3, 3rd \& 6 at the CAR 42-yard line.
Pregame line: CHI favored by 3. Empirical rate $\hat{p}=0.80$; CHI won.}

\begin{tabular}{@{}p{0.085\textwidth}@{\hskip 6pt}p{0.80\textwidth}@{\hskip 6pt}r@{}}
\textit{Base (CoT)} &
\emph{``[\,\dots] Base Probability (Favored by 3): 60--70\%. Adjustment for Poor Performance:
$-10$ to $-20\%$. Final Probability: 40--50\%.''} \ (a leading, favored team, with the stated
value at the bottom of its own adjusted range) & $0.40$ \\[3pt]
\textit{Masked-CoT (ours)} &
\emph{``[\,\dots] CHI is in a strong position to maintain their lead, as they are favored and
have the ball [\,\dots] ample time to either score a touchdown or at least a field goal.
Probability: 85\%.''} & $0.85$ \\
\end{tabular}

\end{minipage}
\caption{Reasoning before and after gradient-masked CoT training, on identical game states
(blinded-judge sample, 2023). In the base completions the stated probability does not follow from
the text: an addition that sums to 108\% reports one of its branches, and a leading favorite is
rated below an even chance. The masked-CoT model reads the live state and states a probability
consistent with it, close to the empirical rate $\hat{p}$. Inconsistency of this kind falls from
22.4\% of base completions to 4.4\% after masked training (\cref{fig:faithfulness}).}
\label{fig:reasoning-examples}
\end{figure*}

%% file: sections/05_setup.tex
\section{Experimental setup}
\label{sec:setup}

States are drawn from National Football League regular-season play-by-play for the 2015 through 2024
seasons, using the public nflfastR data \citep{yurko2018, nflfastr}. Each play is one training
example: the game state of \cref{sec:setting}, paired with the realized outcome of whether the team in
possession won. We split by season into $40{,}246$ training states for 2015 through 2022,
$5{,}241$ for selection in 2023, and $5{,}185$ for test in 2024, so that no game appears in two splits.
The empirical-rate target is estimated only from training-season outcomes, and no prompt contains the
live market win probability; we verified that it appears in none of the $40{,}246$ training prompts.

The base model is Qwen2.5-7B-Instruct, a dense non-thinking model, used with no supervised fine-tuning;
the only training is reinforcement learning against the reward of \cref{eq:reward}. We use group-relative
policy optimization \citep{grpo} as implemented in the TRL library, with vLLM rollouts colocated in the
training process, on a single NVIDIA L40S (48\,GB) GPU. The policy is adapted with LoRA (rank $16$,
$\alpha=32$, dropout $0.05$) on the attention and feed-forward projections, in bfloat16; at this scale one
card holds the policy, the reference model, and the rollout engine without quantization. Each step
samples eight completions per state at temperature $0.9$, uses a token-level loss \citep{dapo}, and
performs a single on-policy update. We disable TRL's vLLM importance-sampling correction, which otherwise
masks the loss under the small per-token mismatch between vLLM and training log-probabilities; with
single-pass updates this recovers standard group-relative optimization. Training runs for $250$ steps
with a $20$-step warmup, saving every $50$; we select the checkpoint by Brier score on the full 2023
split and report it on 2024.

The two models differ as in \cref{sec:decoupling}: the direct model emits the probability in at most $48$
tokens, while the masked model emits up to $640$ tokens of reasoning and applies the gradient only to the
final probability span. Both accumulate gradients over sixteen micro-batches and keep reward scaling on.
The direct model uses a learning rate of $2\times10^{-5}$ and a KL coefficient of $0.01$; the masked model
uses $3\times10^{-5}$ and a KL coefficient of zero. These values come from a learning-rate, batch-size,
and KL sweep reported in \cref{app:hparams}.

We compare against the untrained Qwen2.5-7B-Instruct, prompted with and without a chain of thought; a
frontier model, DeepSeek-V4, prompted zero-shot through its API; the empirical-rate teacher $\hat p$; the
nflverse win-probability model and a gradient-boosted model on the standard feature set
(\cref{sec:results}); and the betting market, converted from odds to a win probability
\citep{strumbelj2014} and used as a near-ceiling reference. We report the Brier score, the expected and
maximum calibration error over ten equal-width bins, accuracy against the realized winner, and the Murphy
decomposition. Confidence intervals are nonparametric bootstraps over plays with $10^4$ resamples, and
two forecasters are compared on the same plays with a paired bootstrap of the per-play Brier difference.
Reliability diagrams use the same ten bins. Checkpoint selection during training uses a fixed random
sample of $128$ held-out 2023 states, scored greedily at each save point; all reported numbers use the
full splits.

%% file: sections/06_results.tex
\section{Results}
\label{sec:results}

\begin{figure*}[t]
\centering
\includegraphics[width=\linewidth]{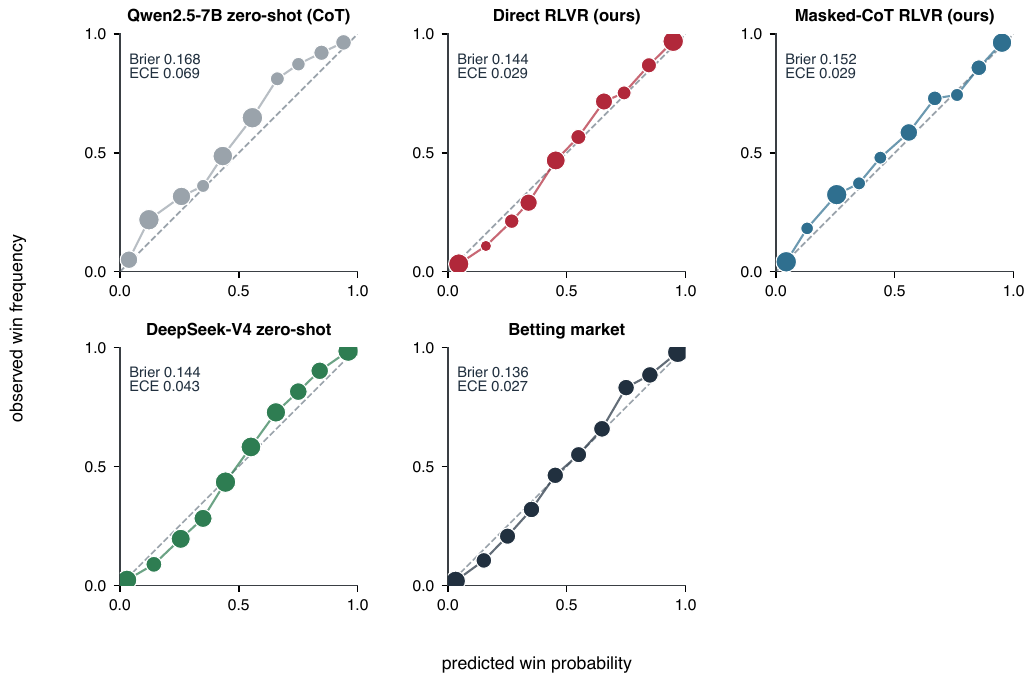}
\caption{Reliability on held-out 2024; points on the diagonal are perfectly calibrated and their area is
proportional to the plays in the bin. The trained forecasters follow the diagonal as closely as the
market, while the untrained chain-of-thought model is overconfident.}
\label{fig:reliability}
\end{figure*}

Reinforcement learning against the conditional-rate reward calibrates the forecaster to the level of
the betting market. On the held-out 2024 season the direct model reaches an ECE of
$0.029$, against the market's $0.027$ and the untrained model's $0.057$. Its maximum calibration error,
$0.060$, is below the market's $0.082$, and its reliability curve lies on the diagonal where the
untrained model's does not (\cref{fig:reliability}). Its Brier score of $0.144$ improves on the untrained
$0.206$ and trails the market's $0.136$ (\cref{tab:main}). The masked chain-of-thought model is as well
calibrated, at an ECE of $0.030$, with a slightly higher Brier score of $0.152$.

\begin{table*}[t]
\centering
\input{tables/T1_main.tex}
\caption{Held-out 2024 ($n=5185$). Brier with $95\%$ bootstrap interval, expected and maximum
calibration error, accuracy, and Murphy resolution. The trained models reach the teacher and
frontier-model tier, the direct model matches the market on calibration, and both trail the market on
Brier through resolution rather than reliability.}
\label{tab:main}
\end{table*}

The direct model matches the market on calibration but trails it on Brier by $0.008$. By the Murphy
decomposition this gap is resolution, not reliability: the model is as well calibrated as the market but
less sharp, $0.106$ against $0.115$. It assigns the right probability to the states it can describe, but
separates fewer of them. The missing sharpness comes from information the prompt does not contain, and
no static forecaster does better.

\begin{figure*}[t]
\centering
\includegraphics[width=\linewidth]{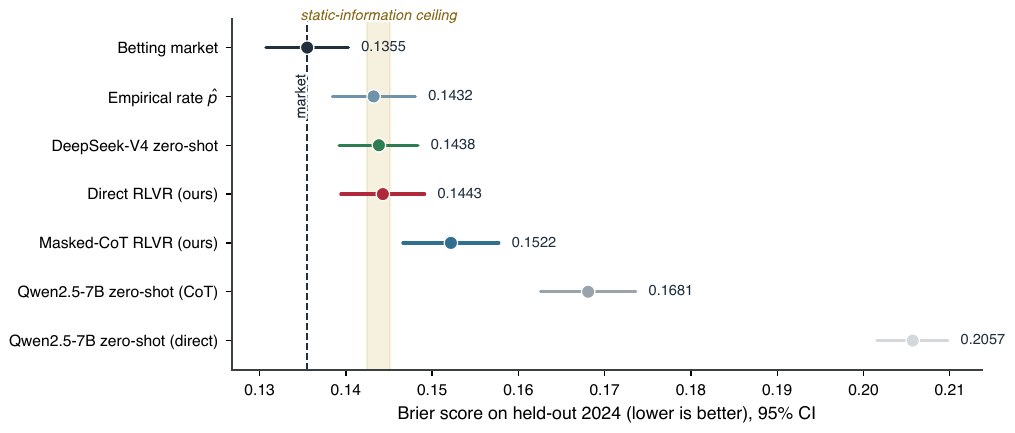}
\caption{Held-out 2024 Brier with $95\%$ bootstrap intervals. A label-free 7B model trained by direct
prediction, a frontier model prompted zero-shot, and the empirical-rate teacher converge near $0.143$
and trail the market by the same significant margin, placing the residual gap in the available
information rather than in model capacity.}
\label{fig:triangulation}
\end{figure*}

The frontier model and the teacher reach the same Brier score as the direct model, $0.144$ and $0.143$
against $0.144$, and a paired bootstrap separates none of the three. All three trail the market by
$0.008$, significant in each case (\cref{fig:triangulation}), and the direct model is better calibrated
than the frontier model, $0.029$ against $0.043$. No richer static model improves on the coarse rate. The
tuned nflverse model and a gradient-boosted model on the full feature set score $0.156$ and $0.158$
against $0.143$ (\cref{tab:ceiling}), since the extra features add play-level noise without game-level
signal. \citet{brill2024} reach the same bound analytically, showing that the correlated structure of
play-by-play data limits the accuracy of any such model. The market's $0.008$ edge is therefore live
in-game information that no static input contains.

\begin{table}[t]
\centering
\input{tables/T4_ceiling.tex}
\caption{Held-out 2024 Brier and ECE for static-feature forecasters. Two models with more features than
the coarse buckets are worse on Brier; the coarse empirical rate is the best static estimator, and the
direct model reaches its level.}
\label{tab:ceiling}
\end{table}

On the test season the direct model is better calibrated than the teacher it was trained on. Its
ECE is $0.029$ against the teacher's $0.044$, and its maximum calibration error $0.060$
against $0.099$, at the same Brier score. The policy smooths the bucketed target into a calibrated
function of the state.

The decalibration of full-completion chain-of-thought training (\cref{sec:decoupling}) does not come
from misreading the game state. A blinded judge scored five state-reading errors (possession, score,
spread, clock, and pregame anchoring) and found each in one to ten percent of completions across the
base, full-completion, and masked models, so all of them read the game correctly. The full-completion
models fail instead at the step from those facts to a number. They write pseudo-quantitative arguments whose probabilities exceed one and
report a figure their own text does not support (\cref{fig:reasoning-examples}). The error is an
inconsistency between the reasoning and the answer (\cref{fig:faithfulness}), not a factual mistake.
Masking the gradient leaves the reading of the state untouched and keeps the number consistent with it,
which restores calibration.

%% file: tables/T1_main.tex
\begin{tabular}{lccccc}
\toprule
System & Brier & ECE & MCE & Acc. & Resolution \\
\midrule
Qwen2.5-7B zero-shot (direct) & 0.2057\,{\scriptsize[0.2016,\,0.2098]} & 0.0569 & 0.2539 & 0.644 & 0.0388 \\
Qwen2.5-7B zero-shot (CoT) & 0.1681\,{\scriptsize[0.1626,\,0.1736]} & 0.0687 & 0.1493 & 0.756 & 0.0875 \\
Masked-CoT RLVR (ours) & 0.1522\,{\scriptsize[0.1466,\,0.1577]} & 0.0293 & 0.0684 & 0.777 & 0.0986 \\
Direct RLVR (ours) & 0.1443\,{\scriptsize[0.1394,\,0.1491]} & 0.0292 & 0.0596 & 0.784 & 0.1058 \\
DeepSeek-V4 zero-shot & 0.1438\,{\scriptsize[0.1392,\,0.1483]} & 0.0430 & 0.0716 & 0.790 & 0.1078 \\
\midrule
Empirical rate $\hat{p}$ & 0.1432\,{\scriptsize[0.1384,\,0.1480]} & 0.0437 & 0.0993 & 0.783 & 0.1083 \\
Betting market & 0.1355\,{\scriptsize[0.1307,\,0.1403]} & 0.0273 & 0.0824 & 0.799 & 0.1148 \\
\bottomrule
\end{tabular}

%% file: tables/T4_ceiling.tex
\begin{tabular}{lcc}
\toprule
Static-feature forecaster & Brier & ECE \\
\midrule
Coarse $\hat{p}$ (our reward) & 0.1432 & 0.0437 \\
nflverse WP model & 0.1562 & 0.0188 \\
GBM, all features & 0.1584 & 0.0260 \\
\midrule
Direct RLVR (ours) & 0.1443 & 0.0292 \\
Betting market & 0.1355 & 0.0273 \\
\bottomrule
\end{tabular}

%% file: sections/09_conclusion.tex
\section{Conclusion}
\label{sec:conclusion}

We have shown that reinforcement learning can calibrate a probabilistic forecaster to the level of the
betting market using only realized game outcomes, without human labels or supervised fine-tuning.
Verifiable-reward training has used the realized label as the reward, which is well behaved when that
label is deterministic but injects irreducible variance when the outcome is stochastic. We reward an
aggregate of those same labels, a state-conditioned empirical rate, which is equally verifiable and far
less noisy, and which turns a proper scoring rule into a usable training signal for events that have no
correct answer. This carries verifiable-reward training from tasks with a right answer to forecasting,
where the answer is itself a probability.

Even with the denoised target, taking the gradient through a chain of thought decalibrates the model,
because optimizing the final probability rewrites the reasoning that produced it. This refines earlier
accounts: the accuracy-calibration gradient conflict \citep{dcpo} and the overconfidence attributed to
the group-standard-deviation normalization of the advantage \citep{uncalibrated_reasoning} both take the
realized outcome as the target, whereas with a denoised target that normalization is instead necessary
and the problem that remains is the gradient's reach into the reasoning. Confining the gradient to the answer
removes it, by dropping the reasoning or by masking the gradient on it, and the masked variant keeps a
chain of thought from which the forecast follows.

Three unrelated estimators converge on the same Brier score: a reinforcement-trained small model, a
frontier model, and a tabular rate. The convergence is strong evidence that the score is the limit set
by the information in the public game state, not a property of any one of them. It is also a practical
test of whether a forecaster has exhausted its inputs: once unrelated methods agree, further accuracy
must come from new information, not a larger model. A market, where one exists, supplies the
independent reference that makes the test sharp.

The method reaches a strong reference only where the public state already carries most of the predictive
signal, and where outcomes are dense and resolved quickly enough to estimate a reliable empirical rate;
sparse or long-horizon events would weaken both the reward and the comparison against a market. Within
these limits the empirical rate could be enriched with more state features or replaced by a learned rate
model to raise the ceiling it imposes, and the same reward and gradient mask should extend to other
aleatoric domains where calibrated probabilities matter and outcomes are eventually observed, from
weather to elections to clinical prognosis. Calibrated forecasting can be learned from outcomes alone,
and a small open model can match a market once its reward and its gradient are aligned with the quantity
it is asked to forecast, while still exposing the reasoning behind each prediction.

%% file: sections/10_appendix.tex

\section{The empirical-rate target}
\label{app:teacher}

The reward target $\hat p(x)$ is a state-conditioned empirical win rate estimated from
training-season outcomes. Each play is placed in a bucket defined by three game-level features: the
score margin of the team in possession, in fourteen bins with edges at $\pm1$, $\pm4$, $\pm7$,
$\pm10$, $\pm14$, and $\pm21$ points; the time remaining, in seven bins with edges at $2$, $5$,
$10$, $15$, $30$, and $45$ minutes; and the public pregame point spread, in nine bins with edges at
$\pm0.5$, $\pm3$, $\pm7$, and $\pm10$ points. A bucket's raw value is the fraction of its training
plays whose possession team won. Sparse buckets are shrunk toward their coarser parents by a
hierarchical empirical-Bayes rule: writing $n$ and $w$ for the count and wins in a bucket and
$\hat p_{\text{parent}}$ for its parent estimate, the bucket value is
$(w + M\,\hat p_{\text{parent}})/(n + M)$ with pseudocount $M=25$, applied from the global rate
downward. The estimate for an evaluation play is read from the training-season table alone, so no
test outcome enters its own target.

Finer state features do not raise the target. A fine table that adds field position and down under
the same hierarchical backoff matches the coarse table on game-level calibration: on the 2023 split
the coarse target scores $0.153$ Brier and $0.012$ ECE against the fine table's $0.153$ and $0.010$,
both close to the market's $0.150$ and $0.019$ (\cref{tab:granularity}). Field position and down
shape the current drive rather than the game outcome, so the reward uses the coarse
score-by-time-by-spread table.

\begin{table}[h]
\centering
\small
\begin{tabular}{lcc}
\toprule
Target (eval 2023) & Brier & ECE \\
\midrule
Coarse: score $\times$ time $\times$ spread & 0.1532 & 0.0117 \\
Fine: $+$ field position $+$ down & 0.1534 & 0.0099 \\
Betting market & 0.1495 & 0.0193 \\
\bottomrule
\end{tabular}
\caption{Teacher-granularity ablation on held-out 2023. Adding field position and down to the
coarse buckets does not improve game-level calibration, so the reward target stays coarse.}
\label{tab:granularity}
\end{table}

\section{Training and hyperparameter selection}
\label{app:hparams}

Every checkpoint is selected by Brier score on a fixed random sample of $128$ held-out 2023 states,
scored greedily at each $50$-step save, and the two models are tuned separately. The numbers below
are this in-training selection metric, with one setting varied at a time around a reference
configuration; the reported test numbers use the full splits.

For the direct model the learning rate matters most. Raising it from $5\times10^{-6}$ to
$1\times10^{-5}$ lowers the in-training Brier from $0.160$ to $0.154$ at equal calibration error, and
$2\times10^{-5}$ sharpens further at a higher calibration error; on the larger batch this higher rate
is the optimum. Effective batch size, set by gradient accumulation, improves Brier up to four games per
step and then saturates. Three negatives are decisive (\cref{tab:direct-sweep}): turning off reward
scaling collapses resolution, blending the realized outcome back into the target at $\lambda=0.5$
reintroduces its variance, and lowering the KL coefficient below $0.01$ worsens the calibration
error. Temperature is neutral between $0.7$ and $1.0$, and the selected configuration varies by
$0.004$ Brier across three seeds. The reported direct model uses a learning rate of $2\times10^{-5}$,
gradient accumulation over sixteen micro-batches, a KL coefficient of $0.01$, and reward scaling on.

\begin{table}[h]
\centering
\small
\begin{tabular}{llcc}
\toprule
Lever & Setting & Brier & ECE \\
\midrule
Reference & lr $1{\times}10^{-5}$, 2 games/step, rate target & 0.154 & 0.050 \\
\midrule
\multirow{2}{*}{Learning rate} & $5\times10^{-6}$ & 0.160 & 0.052 \\
                               & $2\times10^{-5}$ & 0.152 & 0.079 \\
\midrule
\multirow{2}{*}{Effective batch} & 4 games/step (grad.\ accum.\ 16) & 0.149 & 0.099 \\
                                 & 8 games/step (grad.\ accum.\ 32) & 0.152 & 0.100 \\
\midrule
\multirow{2}{*}{Reward target} & realized outcome $y$ & 0.166 & 0.065 \\
                               & blend $\tfrac12 y + \tfrac12\hat p$ & 0.181 & 0.121 \\
\midrule
KL coefficient & $0.003$ & 0.157 & 0.100 \\
Reward scaling & off & 0.178 & 0.064 \\
Temperature & $1.0$ & 0.147 & 0.090 \\
\bottomrule
\end{tabular}
\caption{Direct-model ablation, in-training selection metric (greedy, fixed $n=128$ held-out 2023,
best-Brier checkpoint). Each row changes one setting from the reference. The learning rate and the
effective batch size trade sharpness against calibration; reward scaling, the rate target, and a
non-zero KL coefficient are decisive negatives when removed.}
\label{tab:direct-sweep}
\end{table}

The masked model differs in two settings. The KL anchor must be removed: with the gradient
concentrated on the few answer tokens, a coefficient of $0.01$ makes the $k_3$ KL estimator overflow
and the loss diverge, and $1\times10^{-3}$ degrades calibration over training, while $\beta=0$ is
stable because the answer-span mask already protects the reasoning the reference model would
otherwise anchor. The learning rate follows an inverted-U with a higher optimum than for the direct
model (\cref{tab:masked-sweep}): held-out Brier improves from $5\times10^{-6}$ to a peak at
$3\times10^{-5}$ and collapses at $4\times10^{-5}$, where the unanchored update drives probabilities
to extremes. The reported masked model uses a learning rate of $3\times10^{-5}$ and $\beta=0$.

\begin{table}[h]
\centering
\small
\begin{tabular}{llc}
\toprule
Lever & Setting & Brier \\
\midrule
\multirow{3}{*}{KL coefficient} & $\beta=0$ & 0.174 \\
                                & $\beta=1\times10^{-3}$ & 0.190 \\
                                & $\beta=0.01$ & diverges \\
\midrule
\multirow{5}{*}{Learning rate ($\beta=0$)} & $5\times10^{-6}$ & 0.1745 \\
                                           & $1\times10^{-5}$ & 0.1740 \\
                                           & $2\times10^{-5}$ & 0.1717 \\
                                           & $3\times10^{-5}$ & 0.1708 \\
                                           & $4\times10^{-5}$ & over-sharpens \\
\bottomrule
\end{tabular}
\caption{Masked-model ablation, held-out 2024 (n${=}1000$ subset, best checkpoint). The KL anchor
must be removed, and the learning rate follows an inverted-U with an optimum at $3\times10^{-5}$,
above which the unanchored, concentrated gradient drives the maximum calibration error past $0.5$.}
\label{tab:masked-sweep}
\end{table}

\section{Blinded-judge protocol for reasoning quality}
\label{app:judge}

To measure reasoning quality we collect, for each model, its chain of thought on $250$ held-out 2023
plays stratified across game states, with the ground-truth game facts taken from each play's
features. A separate model judges every completion blind to its source and decides, with a quote
from the completion supporting each judgment, whether the chain of thought misreads any of five game
facts (possession, score, spread, clock, and the pregame favorite) and whether the stated
probability follows from the reasoning. We audited the judge by reading every flagged case; each
flag carries a specific checkable reason, and the judge flags the masked model alongside the others.

State-reading errors are low and flat across the base model, the two full-completion reinforcement
runs, and the masked model. Possession and score are misread in about one percent of completions,
the pregame favorite in two to seven percent, the spread in three to six percent, and the clock in
seven to ten percent, the last a base-model weakness that no training variant changes. Reinforcement
learning does not corrupt the reading of the game.

The models separate instead on whether the stated probability follows from the reasoning
(\cref{tab:faithfulness}). The base model is inconsistent on $22.4\%$ of completions, and
full-completion reinforcement learning does not improve it; the failures are broken probability
arithmetic, drive-level and game-level probabilities conflated, and a final number the text does not
support. The masked model is inconsistent on $4.4\%$. The masked prompt format accounts for much of
the reduction, since a base model read under the masked prompt is inconsistent on $6.8\%$, and
masked training lowers it further. At $250$ plays the format and training intervals overlap on this
checkpoint, while a separate control on the reported checkpoint isolates a significant training
effect.

\begin{table}[h]
\centering
\small
\begin{tabular}{lc}
\toprule
Model & Inconsistent reasoning [95\% CI] \\
\midrule
Base (chain of thought) & 22.4\% \ci{17.6}{27.6} \\
Full-completion RL (run A) & 18.8\% \ci{14.4}{23.6} \\
Full-completion RL (run B) & 14.0\% \ci{10.0}{18.4} \\
Base, masked prompt (no training) & 6.8\% \ci{4.0}{9.6} \\
Masked-CoT RLVR (ours) & 4.4\% \ci{2.0}{7.2} \\
\bottomrule
\end{tabular}
\caption{Rate at which the stated probability does not follow from the chain of thought, blinded
judge over $250$ held-out 2023 plays. Full-completion reinforcement learning does not improve on the
base rate; the masked prompt format reduces it sharply and masked training reduces it further.}
\label{tab:faithfulness}
\end{table}

\section{Data-integrity audit}
\label{app:integrity}

The prompt exposes one market-derived quantity, the public pregame point spread, and never the live
in-game win probability that serves as the evaluation reference. We scanned all $40{,}246$ training
prompts and the 2023 and 2024 evaluation prompts for the market win probability and found no
occurrence. The empirical-rate target is built only from realized outcomes and the pregame spread,
and the target for an evaluation play is computed from the training-season table alone, so neither
the market probability nor any test outcome enters training. The splits are disjoint by season,
training on 2015 through 2022, selecting on 2023, and testing on 2024, so no game appears in more
than one split. Every reported number is recomputed from the per-play predictions by a single
deterministic script, so the tables and figures reproduce from the released predictions without a
GPU.

%% file: main.bbl
\begin{thebibliography}{38}
\providecommand{\natexlab}[1]{#1}
\providecommand{\url}[1]{\texttt{#1}}
\expandafter\ifx\csname urlstyle\endcsname\relax
  \providecommand{\doi}[1]{doi: #1}\else
  \providecommand{\doi}{doi: \begingroup \urlstyle{rm}\Url}\fi

\bibitem[Shao et~al.(2024)Shao, Wang, Zhu, Xu, Song, Bi, Zhang, Zhang, Li, Wu,
  and Guo]{grpo}
Zhihong Shao, Peiyi Wang, Qihao Zhu, Runxin Xu, Junxiao Song, Xiao Bi, Haowei
  Zhang, Mingchuan Zhang, Y.~K. Li, Y.~Wu, and Daya Guo.
\newblock {DeepSeekMath}: Pushing the limits of mathematical reasoning in open
  language models.
\newblock \emph{arXiv preprint arXiv:2402.03300}, 2024.

\bibitem[Guo et~al.(2025)Guo, Yang, Zhang, Song, et~al.]{deepseek_r1}
Daya Guo, Dejian Yang, Haowei Zhang, Junxiao Song, et~al.
\newblock {DeepSeek-R1} incentivizes reasoning in {LLMs} through reinforcement
  learning.
\newblock \emph{Nature}, 645\penalty0 (8081):\penalty0 633--638, 2025.
\newblock arXiv:2501.12948.

\bibitem[Brier(1950)]{brier1950}
Glenn~W. Brier.
\newblock Verification of forecasts expressed in terms of probability.
\newblock \emph{Monthly Weather Review}, 78\penalty0 (1):\penalty0 1--3, 1950.

\bibitem[Gneiting and Raftery(2007)]{gneiting2007}
Tilmann Gneiting and Adrian~E. Raftery.
\newblock Strictly proper scoring rules, prediction, and estimation.
\newblock \emph{Journal of the American Statistical Association}, 102\penalty0
  (477):\penalty0 359--378, 2007.

\bibitem[Leng et~al.(2024)Leng, Huang, Zhu, and Huang]{leng2024}
Jixuan Leng, Chengsong Huang, Banghua Zhu, and Jiaxin Huang.
\newblock Taming overconfidence in {LLMs}: Reward calibration in {RLHF}.
\newblock \emph{arXiv preprint arXiv:2410.09724}, 2024.

\bibitem[Ma et~al.(2026)Ma, Wen, Cao, Lu, Lin, Yang, He, Han, and Sun]{dcpo}
Zhengzhao Ma, Xueru Wen, Boxi Cao, Yaojie Lu, Hongyu Lin, Jinglin Yang, Min He,
  Xianpei Han, and Le~Sun.
\newblock Decoupling reasoning and confidence: Resurrecting calibration in
  reinforcement learning from verifiable rewards.
\newblock In \emph{Proceedings of the 43rd International Conference on Machine
  Learning (ICML)}, 2026.
\newblock arXiv:2603.09117.

\bibitem[Bereket and Leskovec(2025)]{uncalibrated_reasoning}
Michael Bereket and Jure Leskovec.
\newblock Uncalibrated reasoning: {GRPO} induces overconfidence for stochastic
  outcomes.
\newblock \emph{arXiv preprint arXiv:2508.11800}, 2025.

\bibitem[Damani et~al.(2025)Damani, Puri, Slocum, Shenfeld, Choshen, Kim, and
  Andreas]{rlcr}
Mehul Damani, Isha Puri, Stewart Slocum, Idan Shenfeld, Leshem Choshen, Yoon
  Kim, and Jacob Andreas.
\newblock Beyond binary rewards: Training {LMs} to reason about their
  uncertainty.
\newblock \emph{arXiv preprint arXiv:2507.16806}, 2025.

\bibitem[Bani-Harouni et~al.(2025)Bani-Harouni, Pellegrini, Stangel, {\"O}zsoy,
  Zaripova, Navab, and Keicher]{rewarding_doubt}
David Bani-Harouni, Chantal Pellegrini, Paul Stangel, Ege {\"O}zsoy, Kamilia
  Zaripova, Nassir Navab, and Matthias Keicher.
\newblock Rewarding doubt: A reinforcement learning approach to calibrated
  confidence expression of large language models.
\newblock \emph{arXiv preprint arXiv:2503.02623}, 2025.

\bibitem[Band et~al.(2024)Band, Li, Ma, and Hashimoto]{band2024}
Neil Band, Xuechen Li, Tengyu Ma, and Tatsunori Hashimoto.
\newblock Linguistic calibration of long-form generations.
\newblock In \emph{Proceedings of the 41st International Conference on Machine
  Learning (ICML)}, 2024.
\newblock arXiv:2404.00474.

\bibitem[Halawi et~al.(2024)Halawi, Zhang, Yueh-Han, and
  Steinhardt]{halawi2024}
Danny Halawi, Fred Zhang, Chen Yueh-Han, and Jacob Steinhardt.
\newblock Approaching human-level forecasting with language models.
\newblock In \emph{Advances in Neural Information Processing Systems
  (NeurIPS)}, 2024.
\newblock arXiv:2402.18563.

\bibitem[Pratt et~al.(2024)Pratt, Blumberg, Carolino, and Morris]{pratt2024}
Sarah Pratt, Seth Blumberg, Pietro~Kreitlon Carolino, and Meredith~Ringel
  Morris.
\newblock Can language models use forecasting strategies?
\newblock \emph{arXiv preprint arXiv:2406.04446}, 2024.

\bibitem[Turtel et~al.(2025{\natexlab{a}})Turtel, Franklin, Skotheim, Hewitt,
  and Schoenegger]{turtel2025outcome}
Benjamin Turtel, Danny Franklin, Kris Skotheim, Luke Hewitt, and Philipp
  Schoenegger.
\newblock Outcome-based reinforcement learning to predict the future.
\newblock \emph{arXiv preprint arXiv:2505.17989}, 2025{\natexlab{a}}.

\bibitem[Turtel et~al.(2025{\natexlab{b}})Turtel, Franklin, and
  Schoenegger]{turtel2025selftaught}
Benjamin Turtel, Danny Franklin, and Philipp Schoenegger.
\newblock {LLMs} can teach themselves to better predict the future.
\newblock \emph{arXiv preprint arXiv:2502.05253}, 2025{\natexlab{b}}.

\bibitem[Paleka et~al.(2025)Paleka, Goel, Geiping, and Tram{\`e}r]{paleka2025}
Daniel Paleka, Shashwat Goel, Jonas Geiping, and Florian Tram{\`e}r.
\newblock Pitfalls in evaluating language model forecasters.
\newblock \emph{arXiv preprint arXiv:2506.00723}, 2025.

\bibitem[Karger et~al.(2024)Karger, Bastani, Yueh-Han, Jacobs, Halawi, Zhang,
  and Tetlock]{karger2024forecastbench}
Ezra Karger, Houtan Bastani, Chen Yueh-Han, Zachary Jacobs, Danny Halawi, Fred
  Zhang, and Philip~E. Tetlock.
\newblock {ForecastBench}: A dynamic benchmark of {AI} forecasting
  capabilities.
\newblock \emph{arXiv preprint arXiv:2409.19839}, 2024.

\bibitem[Yu et~al.(2025)Yu, Zhang, Zhu, Yuan, Zuo, Yue, et~al.]{dapo}
Qiying Yu, Zheng Zhang, Ruofei Zhu, Yufeng Yuan, Xiaochen Zuo, Yu~Yue, et~al.
\newblock {DAPO}: An open-source {LLM} reinforcement learning system at scale.
\newblock \emph{arXiv preprint arXiv:2503.14476}, 2025.

\bibitem[Chu et~al.(2026)Chu, Huang, Zhang, Wei, and Wang]{gpg2026}
Xiangxiang Chu, Hailang Huang, Xiao Zhang, Fei Wei, and Yong Wang.
\newblock {GPG}: A simple and strong reinforcement learning baseline for model
  reasoning.
\newblock In \emph{International Conference on Learning Representations
  (ICLR)}, 2026.
\newblock arXiv:2504.02546.

\bibitem[Zhang et~al.(2026)Zhang, Liu, Yuan, Yuan, Gu, and Yao]{rpg2026}
Yifan Zhang, Yifeng Liu, Huizhuo Yuan, Yang Yuan, Quanquan Gu, and Andrew
  Chi-Chih Yao.
\newblock On the design of {KL}-regularized policy gradient algorithms for
  {LLM} reasoning.
\newblock In \emph{International Conference on Learning Representations
  (ICLR)}, 2026.
\newblock arXiv:2505.17508.

\bibitem[Wang et~al.(2025{\natexlab{a}})Wang, Yu, Gao, Zheng, Liu, Lu, Dang,
  Chen, Yang, Zhang, Liu, Yang, Zhao, Yue, Song, Yu, Huang, and
  Lin]{wang2025entropy}
Shenzhi Wang, Le~Yu, Chang Gao, Chujie Zheng, Shixuan Liu, Rui Lu, Kai Dang,
  Xionghui Chen, Jianxin Yang, Zhenru Zhang, Yuqiong Liu, An~Yang, Andrew Zhao,
  Yang Yue, Shiji Song, Bowen Yu, Gao Huang, and Junyang Lin.
\newblock Beyond the 80/20 rule: High-entropy minority tokens drive effective
  reinforcement learning for {LLM} reasoning.
\newblock In \emph{Advances in Neural Information Processing Systems
  (NeurIPS)}, 2025{\natexlab{a}}.
\newblock arXiv:2506.01939.

\bibitem[Wang et~al.(2025{\natexlab{b}})Wang, Liu, Zhang, Li, Zhou, and
  Pan]{archer2025}
Jiakang Wang, Runze Liu, Fuzheng Zhang, Xiu Li, Guorui Zhou, and Ling Pan.
\newblock Stabilizing knowledge, promoting reasoning: Dual-token constraints
  for {RLVR}.
\newblock \emph{arXiv preprint arXiv:2507.15778}, 2025{\natexlab{b}}.

\bibitem[Tan et~al.(2025)Tan, Wang, Pan, Lin, Wang, Wu, Chen, Zheng, Tang, and
  Yang]{gtpo2025}
Hongze Tan, Zihan Wang, Jianfei Pan, Jinghao Lin, Hao Wang, Yifan Wu, Tao Chen,
  Zhihang Zheng, Zhihao Tang, and Haihua Yang.
\newblock {GTPO} and {GRPO-S}: Token and sequence-level reward shaping with
  policy entropy.
\newblock \emph{arXiv preprint arXiv:2508.04349}, 2025.

\bibitem[Lock and Nettleton(2014)]{lock2014}
Dennis Lock and Dan Nettleton.
\newblock Using random forests to estimate win probability before each play of
  an {NFL} game.
\newblock \emph{Journal of Quantitative Analysis in Sports}, 10\penalty0
  (2):\penalty0 197--205, 2014.

\bibitem[Yurko et~al.(2019)Yurko, Ventura, and Horowitz]{yurko2018}
Ronald Yurko, Samuel Ventura, and Maksim Horowitz.
\newblock {nflWAR}: A reproducible method for offensive player evaluation in
  football.
\newblock \emph{Journal of Quantitative Analysis in Sports}, 15\penalty0 (3),
  2019.
\newblock arXiv:1802.00998.

\bibitem[Carl and Baldwin(2024)]{nflfastr}
Sebastian Carl and Ben Baldwin.
\newblock \emph{{nflfastR}: Functions to Efficiently Access {NFL} Play by Play
  Data}, 2024.
\newblock URL \url{https://www.nflfastr.com/}.
\newblock R package.

\bibitem[Brill et~al.(2025)Brill, Yurko, and Wyner]{brill2024}
Ryan~S. Brill, Ronald Yurko, and Abraham~J. Wyner.
\newblock Exploring the difficulty of estimating win probability: A simulation
  study.
\newblock \emph{Journal of Quantitative Analysis in Sports}, 2025.
\newblock arXiv:2406.16171.

\bibitem[Polson and Stern(2015)]{polson2015}
Nicholas~G. Polson and Hal~S. Stern.
\newblock The implied volatility of a sports game.
\newblock \emph{Journal of Quantitative Analysis in Sports}, 11\penalty0
  (3):\penalty0 145--153, 2015.

\bibitem[Boulier and Stekler(2003)]{boulier2003}
Bryan~L. Boulier and H.~O. Stekler.
\newblock Predicting the outcomes of national football league games.
\newblock \emph{International Journal of Forecasting}, 19\penalty0
  (2):\penalty0 257--270, 2003.

\bibitem[Levitt(2004)]{levitt2004}
Steven~D. Levitt.
\newblock Why are gambling markets organised so differently from financial
  markets?
\newblock \emph{The Economic Journal}, 114\penalty0 (495):\penalty0 223--246,
  2004.

\bibitem[Franck et~al.(2010)Franck, Verbeek, and N{\"u}esch]{franck2010}
Egon Franck, Erwin Verbeek, and Stephan N{\"u}esch.
\newblock Prediction accuracy of different market structures: Bookmakers versus
  a betting exchange.
\newblock \emph{International Journal of Forecasting}, 26\penalty0
  (3):\penalty0 448--459, 2010.

\bibitem[Cox et~al.(2021)Cox, Schwartz, Van~Ness, and Van~Ness]{cox2021}
Justin Cox, Adam~L. Schwartz, Bonnie~F. Van~Ness, and Robert~A. Van~Ness.
\newblock The predictive power of college football spreads: Regular season
  versus bowl games.
\newblock \emph{Journal of Sports Economics}, 22\penalty0 (3):\penalty0
  251--273, 2021.

\bibitem[{\v{S}}trumbelj(2014)]{strumbelj2014}
Erik {\v{S}}trumbelj.
\newblock On determining probability forecasts from betting odds.
\newblock \emph{International Journal of Forecasting}, 30\penalty0
  (4):\penalty0 934--943, 2014.

\bibitem[Murphy(1973)]{murphy1973}
Allan~H. Murphy.
\newblock A new vector partition of the probability score.
\newblock \emph{Journal of Applied Meteorology}, 12\penalty0 (4):\penalty0
  595--600, 1973.

\bibitem[Guo et~al.(2017)Guo, Pleiss, Sun, and Weinberger]{guo2017}
Chuan Guo, Geoff Pleiss, Yu~Sun, and Kilian~Q. Weinberger.
\newblock On calibration of modern neural networks.
\newblock In \emph{Proceedings of the 34th International Conference on Machine
  Learning (ICML)}, 2017.
\newblock arXiv:1706.04599.

\bibitem[Br{\"o}cker and Smith(2007)]{brocker2007}
Jochen Br{\"o}cker and Leonard~A. Smith.
\newblock Increasing the reliability of reliability diagrams.
\newblock \emph{Weather and Forecasting}, 22\penalty0 (3):\penalty0 651--661,
  2007.

\bibitem[Dimitriadis et~al.(2021)Dimitriadis, Gneiting, and
  Jordan]{dimitriadis2021}
Timo Dimitriadis, Tilmann Gneiting, and Alexander~I. Jordan.
\newblock Stable reliability diagrams for probabilistic classifiers.
\newblock \emph{Proceedings of the National Academy of Sciences}, 118\penalty0
  (8):\penalty0 e2016191118, 2021.

\bibitem[Efron(1979)]{efron1979}
Bradley Efron.
\newblock Bootstrap methods: Another look at the jackknife.
\newblock \emph{The Annals of Statistics}, 7\penalty0 (1):\penalty0 1--26,
  1979.

\bibitem[Efron and Morris(1975)]{efron_morris1975}
Bradley Efron and Carl Morris.
\newblock Data analysis using {Stein's} estimator and its generalizations.
\newblock \emph{Journal of the American Statistical Association}, 70\penalty0
  (350):\penalty0 311--319, 1975.

\end{thebibliography}
